\documentclass[10pt,twocolumn,letterpaper]{article}

\usepackage{cvpr}              
\usepackage{array}

\usepackage{xcolor}  
\definecolor{cvprblue}{rgb}{0.21,0.49,0.74}
\usepackage[pagebackref,breaklinks,colorlinks,allcolors=cvprblue]{hyperref}

\def\paperID{13968} 
\def\confName{CVPR}
\def\confYear{2026}

\definecolor{softpink}{HTML}{ED028C}
\hypersetup{
    urlcolor=softpink
}

\title{nuCarla: A nuScenes-Style Bird’s-Eye View Perception \\ Dataset for CARLA Simulation}

\author{Zhijie Qiao\textsuperscript{1}, Zhong Cao\textsuperscript{1}, Henry X. Liu\textsuperscript{1,2} \\ [0.5em]
\textsuperscript{1} Civil and Environmental Engineering, University of Michigan \\
\textsuperscript{2} Transportation Research Institute, University of Michigan \\ [0.3em] 
{\normalsize\textcolor{softpink}{\url{https://github.com/michigan-traffic-lab/nuCarla}}}
}

\begin{document}
\maketitle
\begin{abstract}


End-to-end (E2E) autonomous driving heavily relies on closed-loop simulation, where perception, planning, and control are jointly trained and evaluated in interactive environments. Yet, most existing datasets are collected from the real world under non-interactive conditions, primarily supporting open-loop learning while offering limited value for closed-loop testing. Due to the lack of standardized, large-scale, and thoroughly verified datasets to facilitate learning of meaningful intermediate representations, such as bird’s-eye-view (BEV) features, closed-loop E2E models remain far behind even simple rule-based baselines.
To address this challenge, we introduce nuCarla, a large-scale, nuScenes-style BEV perception dataset built within the CARLA simulator. nuCarla features (1) full compatibility with the nuScenes format, enabling seamless transfer of real-world perception models; (2) a dataset scale comparable to nuScenes, but with more balanced class distributions; (3) direct usability for closed-loop simulation deployment; and (4) high-performance BEV backbones that achieve state-of-the-art detection results.
By providing both data and models as open benchmarks, nuCarla substantially accelerates closed-loop E2E development, paving the way toward reliable and safety-aware research in autonomous driving.

\end{abstract}    
\section{Introduction}
\label{sec:intro}

In the field of autonomous driving, end-to-end (E2E) systems have attracted increasing attention. UniAD~\cite{uniad} represents an influential milestone, proposing a transformer-based architecture that unifies perception, prediction, and planning through a query-driven design, achieving state-of-the-art performance on the large-scale nuScenes~\cite{nuscenes} dataset. Subsequent works such as VAD~\cite{vad} and UAD~\cite{uad} further improved the overall architectural scheme and established new records on open-loop prediction tasks. Despite these advances, studies have highlighted that improvements in open-loop do not necessarily translate to better performance in closed-loop evaluation~\cite{zhai2023ADMLP, closingloopmotionprediction}. In open-loop, the ego agent is reset to the ground truth position at each frame, breaking the causal relationship between actions and outcomes. In contrast, closed-loop evaluation enforces continuous control, where minor disturbances can accumulate and cause the agent to drift off the track~\cite{dagger}. 

\begin{figure}[!t]
    \centering
    \vspace{0.15cm}
    \includegraphics[width=\linewidth]{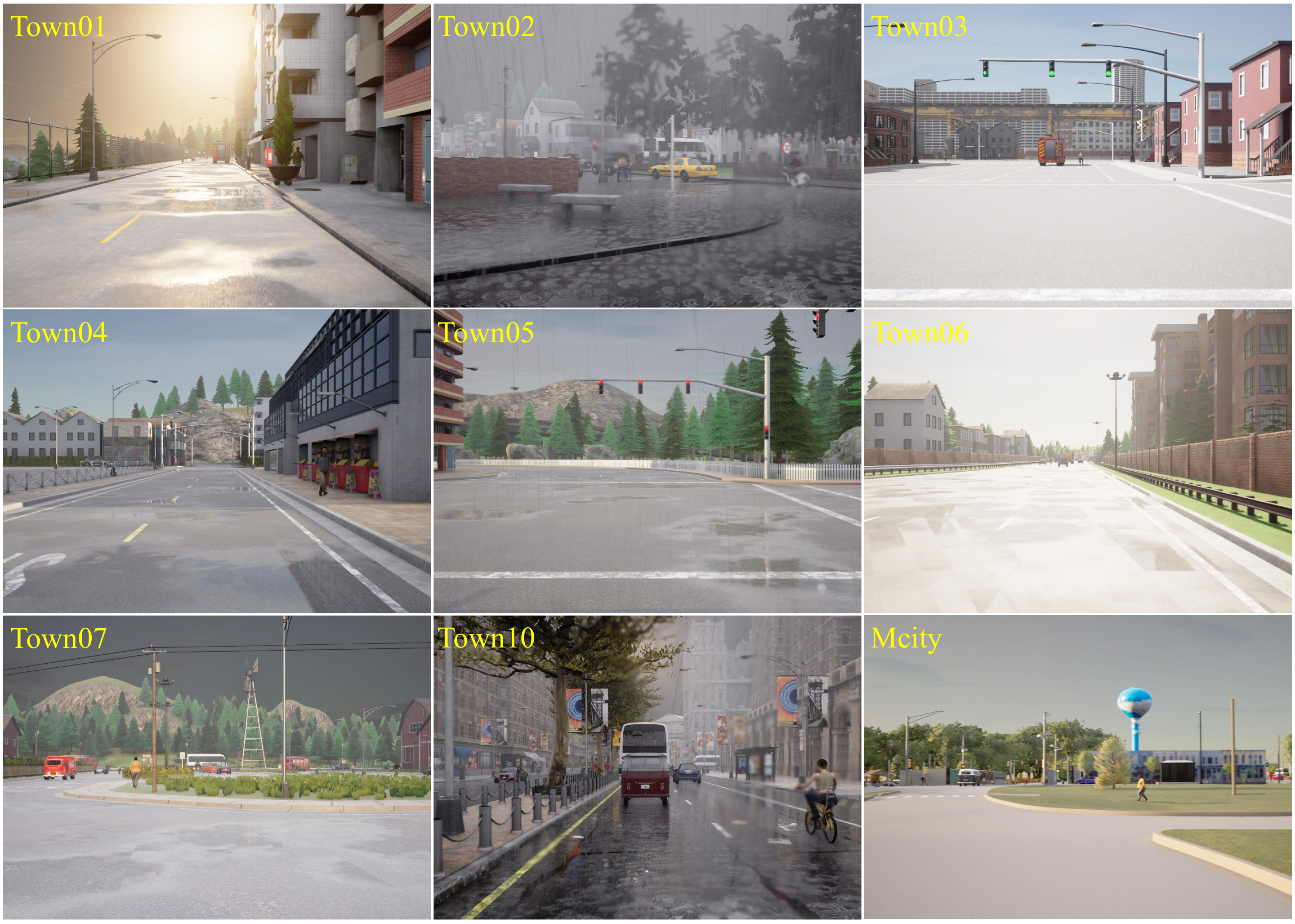}
    \caption{Maps of nine CARLA towns with traffic in the nuCarla dataset, shown under diverse weather conditions.}
    \label{fig:towns}
\end{figure}

Since real-world testing is risky, inefficient, and often non-reproducible, closed-loop training and evaluation for autonomous vehicles are extensively conducted on simulation environments. Although there have been advances in open-loop E2E modeling, primarily driven by non-interactive real-world datasets~\cite{nuscenes, waymo, waymo-e2e}, these developments have not yet been reflected in the simulation domain. To date, there remains a lack of standardized, thoroughly verified, and large-scale datasets specifically designed for simulation-based closed-loop research. 

Among existing platforms, CARLA~\cite{carla} has become widely used for autonomous driving simulation, offering a high-fidelity, physics simulator with diverse sensor suites and realistic environments. The CARLA autonomous driving leaderboard~\cite{leaderboard} provides an open framework for evaluating autonomous agents on predefined routes. Despite its popularity, the leaderboard has been criticized for focusing on basic driving skills and lacking rigorous evaluation under complex conditions~\cite{chen2022learning}. To address this, Bench2Drive~\cite{bench2drive} introduced a large-scale dataset with well-defined metrics and provided pretrained E2E models based on UniAD~\cite{uniad} and VAD~\cite{vad} architectures. However, the reported driving success rates (SR) for these models remain notably low, at only 16.36\% and 15.00\%, respectively.

Subsequent works have improved performance on the benchmark. For example, MomAD~\cite{momad} achieved an SR of 16.71\%, VeteranAD~\cite{veteranad} 33.85\%, DriveTransformer~\cite{drivetransformer} 35.01\%, and Orion~\cite{fu2025orion} 54.62\%. Nevertheless, even the best-performing framework, Orion, which leverages the advanced Vision-Language Model (VLM) Qwen2~\cite{qwen2} for scene understanding and navigation reasoning, succeeds in only about half of the scenarios, each lasting merely 20 seconds. In contrast, a simple rule-based algorithm, PDM-Lite~\cite{pdm-lite}, originally designed for Graph Visual Question Answering tasks, achieves a remarkable 92.27\% SR when utilizing ground-truth perception information~\cite{leaderboardv2}.

We argue that the suboptimal performance of existing models does not necessarily stem from the E2E paradigm itself, but rather from limitations in the available data. \textit{Most simulation datasets provide only raw sensor inputs and direct vehicle control outputs, restricting E2E systems' ability to learn meaningful intermediate representations, such as bird’s-eye-view (BEV) features, which are critical for improved generalization and stability}. Yet, such intermediate-level datasets and pretrained perception backbones remain largely unexplored in the literature, making it difficult to develop robust closed-loop systems.

To address this gap, we introduce nuCarla, a nuScenes-style, camera-based BEV perception dataset built within the CARLA simulator. Following the standard nuScenes protocol, nuCarla contains 1,000 driving scenarios (700 for training, 150 for validation, and 150 for testing), each consisting of 40 frames sampled at 0.5-second intervals. The dataset strictly aligns with nuScenes in naming conventions, annotation structure, file hierarchy, and API compatibility, enabling direct transfer of existing BEV models to the CARLA environment without modification.

To thoroughly validate the nuCarla dataset, we train and evaluate four state-of-the-art BEV perception models: BEVFormer~\cite{bevformer, bevformerv2}, PETR~\cite{petr, petrv2}, BEVDet~\cite{bevdet, bevdet4d}, and FastBEV~\cite{fastbev}. All models achieve stable convergence and demonstrate strong detection performance on the nuCarla validation set, as measured by the official nuScenes metrics. 

We release the full dataset along with pretrained weights for each BEV architecture. In the same way that modern perception frameworks adopt ResNet~\cite{resnet} or VoVNet~\cite{vovnet} as standard visual backbones, we envision nuCarla as a BEV-level perception backbone for the development of robust E2E autonomous driving systems.

Finally, we resolve version conflicts by upgrading the legacy mmdetection3d-1.0~\cite{mmdet3d} frameworks (widely adopted in perception models and E2E systems) to ensure full compatibility with the latest PyTorch and GPU architectures. This resolves a persistent pain point in the research community~\cite{issue313, issue245, issue206}.

The main contributions of this work are as follows:
\begin{enumerate}[leftmargin=2.5em]
    \item We provide a nuScenes-style BEV perception dataset in the CARLA simulator to facilitate development of perception models.
    \item We validate the dataset by training four BEV models, achieving competitive performance under the official nuScenes metrics. We also upgrade the legacy mmdetection3d framework to be compatible with the latest PyTorch and GPU architectures.
    \item We release pretrained weights for all evaluated BEV architectures, providing strong perception backbones to support future E2E autonomous driving research.

\end{enumerate}
\section{Related Work}
\label{sec:related-work}

\subsection{Open-Loop Trajectory Prediction}

Early E2E autonomous driving frameworks are primarily implemented and evaluated on open-loop trajectory prediction tasks. UniAD~\cite{uniad} introduces a unified transformer-based architecture that formulates perception, prediction, and planning as interdependent query-based tasks, thereby mitigating error propagation across subtasks and achieving leading results on multiple benchmarks of the nuScenes~\cite{nuscenes} dataset. VAD~\cite{vad} proposes a vectorized approach that replaces rasterized inputs with instance-level vector representations, modeling agents and map features as explicit geometric entities for improved interpretability and inference speed. GenAD~\cite{genad} is a generative framework that produces driving plans from raw sensor inputs by encoding scenes into instance tokens, learning trajectory priors in a latent space, and modeling agent and ego dynamics. Para-Drive~\cite{paradrive} explores differentiable modular architectures and designs a fully parallel structure that enhances safety and runtime. Hydra-MDP~\cite{hydra} uses knowledge distillation from both human and rule-based experts and generates diverse trajectory candidates through a multi-head decoder that accounts for multiple evaluation metrics; this method achieved first place in the Navsim challenge~\cite{navsim}. Hydra-MDP++~\cite{hydra++} extends Hydra-MDP~\cite{hydra} by introducing richer behavioral evaluation metrics and a lightweight ResNet-34 backbone, achieving additional gains. UAD~\cite{uad} introduces an unsupervised vision-based framework that eliminates manual 3D annotations, using an angular perception pretext to learn spatiotemporal dynamics, achieving greater robustness and efficiency. Despite their architectural differences and methodological advances, these approaches are mainly benchmarked in open-loop settings, breaking causal relationship between decisions and future observations, and may not necessarily lead to good closed-loop driving quality~\cite{zhai2023ADMLP, closingloopmotionprediction}.

\subsection{Closed-Loop Autonomous Driving}

Recognizing the gap between open- and closed-loop evaluations, the research community has increasingly turned to simulation-based platforms to facilitate more comprehensive assessment of E2E systems. Since real-world testing remains expensive and time-consuming for most research groups, CARLA~\cite{carla} has become widely adopted among existing platforms. For example, VAD-v2~\cite{vadv2} extended its open-loop trajectory prediction tasks by conducting closed-loop evaluations on the Town05 benchmark~\cite{leaderboard}. However, the code for its closed-loop development is not publicly available, limiting other researchers’ ability to reproduce or build upon these results. UniAD~\cite{uad} also reported closed-loop testing but does not release its source code.

To address these limitations, Bench2Drive~\cite{bench2drive} released a large-scale CARLA-based dataset designed to accelerate E2E autonomous driving development, providing pretrained models based on the UniAD and VAD architectures. However, these models exhibit poor performance, achieving success rates of only 16.36\% and 15.00\% across 220 evaluation routes, each lasting approximately 20 seconds. Subsequent works report incremental improvements. For example, MomAD (16.71\%)~\cite{momad} introduces momentum to stabilize long-horizon planning by employing topological trajectory matching with Hausdorff distance and cross-attending historical queries. VeteranAD (33.85\%)~\cite{veteranad} integrates perception directly into planning through a “perception-in-plan” design guided by multi-mode trajectory priors and autoregressive updates. DriveTransformer (35.01\%)~\cite{drivetransformer} introduces a task-parallel transformer architecture that enables symmetric interaction among perception, prediction, and planning through sparse queries and streaming updates, improving training stability and scalability. Orion (54.62\%)~\cite{fu2025orion} aligns vision-language reasoning with trajectory generation via a unified framework that combines a QT-Former and a generative planner. 

However, none of these methods match the performance of the simple rule-based planner PDM-Lite~\cite{pdm-lite}, highlighting concerns with current training practices. It also underscores the need to first establish a rigorous perception backbone that effectively captures meaningful intermediate representations, such as bird’s-eye-view (BEV) features, which can then serve as a foundation for further E2E development.

\subsection{BEV Perception Models}

Bird’s-Eye View (BEV) perception models are broadly categorized into three groups: LiDAR-based, camera-based, and sensor fusion. LiDAR-based approaches such as CenterPoint~\cite{centerpoint} detects and tracks 3D objects by focusing on their centers using a keypoint detector, while regressing additional attributes such as size, orientation, and velocity. VoxelNet~\cite{voxelnet} introduces a voxel feature encoding layer that transforms sparse point clouds into unified volumetric features, removing the reliance on manual feature engineering.

Camera-based models are popular due to their lower sensor cost and ability to capture rich semantic information. BEVFormer~\cite{bevformer, bevformerv2} unifies BEV representations via spatiotemporal transformers, aggregating spatial cues from multiple views and temporal information from past frames. PETR~\cite{petr, petrv2} embeds 3D positional information into image features, allowing object queries to access spatial context and enhancing detection accuracy. BEVDet~\cite{bevdet, bevdet4d} incorporates temporal information by fusing features from adjacent frames, which reduces velocity prediction errors. FastBEV~\cite{fastbev} is optimized for efficiency and real-time inference, leveraging lightweight view transformation and multi-frame fusion to deliver both accuracy and speed for on-vehicle deployment.

Sensor fusion approaches combine multiple modalities to improve the accuracy of perception systems. CenterFusion~\cite{centerfusion} introduces a middle-fusion framework that integrates radar and camera through a frustum-based method. BEVDepth~\cite{bevdepth} incorporates explicit LiDAR-based depth supervision and a refinement module to address depth estimation challenges in camera-based detection, resulting in robust inference performance.
\section{Methodology}

In this section, we present the detailed process of constructing the nuCarla dataset, including map selection, weather configuration, traffic generation, ego-vehicle sensor setup, and ground-truth annotation procedures.

\subsection{Maps Selection}

We construct our dataset from nine distinct CARLA maps, including Town01, Town02, Town03, Town04, Town05, Town06, Town07, Town10, and Mcity Ditigal Twin (a high-fidelity simulation of the real-world autonomous vehicle test facility at the University of Michigan)~\cite{mcity}. Together, these maps represent a rich diversity of driving environments, spanning dense urban centers, suburban neighborhoods, and rural roadways. By leveraging the unique characteristics of each town, such as varying road geometries, intersection layouts, and traffic densities, our dataset provides a comprehensive testbed for evaluating perception models under a wide range of realistic scenarios. Town08 and Town09 maps are not included, as they are reserved for the leaderboard challenge and not publicly available. 

In alignment with nuScenes~\cite{nuscenes}, which consists of 1,000 scenarios, our dataset is split into 700 training, 150 validation, and 150 testing scenarios. The 850 training and validation scenarios are evenly distributed across Town01 through Town07, while Town10 and Mcity are reserved for testing in unseen environments.

\subsection{Weather Configuration}

To enhance diversity, we adopt all 14 predefined weather configurations available in CARLA, encompassing various conditions such as sunny, cloudy, and rainy, as well as different times of day including noon and sunset. For each scenario, a random weather condition is applied from this set, resulting in a final distribution that is approximately uniform across all available weather types.

\subsection{Traffic Configuration}

The nuCarla dataset includes six object classes, corresponding to the most safety-critical traffic participants: car, truck, bus, pedestrian, motorcycle, and bicycle. The remaining four classes present in nuScenes, namely construction vehicle, trailer, barrier, and traffic cone, are omitted due to practical constraints. Construction vehicles and trailers are not included because CARLA does not provide the necessary object blueprints. For static objects such as barriers and traffic cones, we have not identified a consistent method for placement across diverse environments.

For the included object classes, we incorporated as many variations as possible to enhance visual and behavioral diversity. Specifically, we included 23 types of cars, 3 types of trucks, 3 types of buses, 46 pedestrian models (38 adults and 8 children), 4 types of motorcycles, and 3 types of bicycles. Currently, we include only actively traveling participants, such as moving vehicles, cycles with riders, and walking pedestrians, while excluding stationary ones such as parked vehicles, cycles without riders, or pedestrians sitting or lying down (note that participants temporarily stopped for traffic lights or yielding are treated as traveling). As the primary objective of this project focuses on perception rather than realistic traffic behavior modeling, we did not adopt advanced traffic control workflows, but instead relied on the default CARLA traffic manager to control all participants.

We further adjusted the distribution of generated traffic participants to create a more balanced dataset. In nuScenes, cars and pedestrians dominate the annotations, while motorcycles and bicycles are severely underrepresented. Although this reflects real-world traffic distributions, such imbalance can cause perception models to perform well on frequent classes but underperform on rare ones. For example, the mean Average Precision (mAP) of BEVFormer~\cite{bevformer} is 0.618 for cars but only 0.398 for bicycles.

\begin{figure}[!t]
    \centering
    \vspace{0.15cm}
    \includegraphics[width=\linewidth]{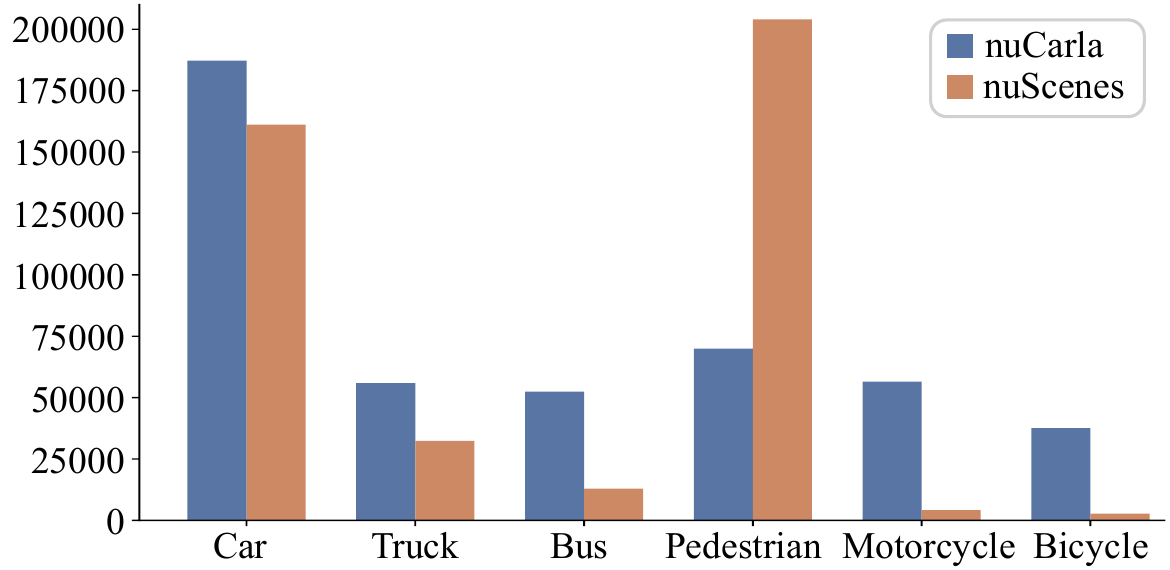}
    \caption{Comparison of class-wise distributions of actively traveling participants in nuCarla and nuScenes across six object classes.}
    \label{fig:distribution}
\end{figure}

We generated 125 participants for each scenario, including 40 cars, 10 trucks, 10 buses, 25 pedestrians, 20 motorcycles, and 20 bicycles. This ensures sufficient traffic density without causing congestion. Given their larger physical sizes, trucks and buses are more visible than smaller participants such as pedestrians, which compensates for their relatively lower occurrence frequencies. In total, this yields 459,632 annotated samples across six object classes. In comparison, the nuScenes dataset contains 417,609 actively traveling participants across the same six classes, making nuCarla comparable in scale but with a more balanced class distribution. Fig.~\ref{fig:distribution} illustrates the class-wise distributions of actively traveling participants in both datasets.



\subsection{Ego and Sensor Configuration}

We selected the Nissan Micra (\(3.63\,\mathrm{m} \times 1.84\,\mathrm{m} \times 1.50\,\mathrm{m}\)) from the available CARLA blueprints to closely match the dimensions of the nuScenes data acquisition vehicle, a Renault Zoe (\(4.08\,\mathrm{m} \times 1.78\,\mathrm{m} \times 1.56\,\mathrm{m}\)). Six RGB cameras (front, front-left, front-right, back, back-left, and back-right) are mounted on the ego vehicle, with an image resolution of \(1600 \times 900\), following the same setup as nuScenes. The intrinsic camera calibration parameters are directly retrieved from CARLA. Note that the dataset does not include any LiDAR or radar sensors. While integrating these sensors is straightforward, this project is primarily focused on camera-based perception. Moreover, introducing insufficiently validated sensor data would risk compromising the reliability of downstream tasks. In future work, we plan to extend the dataset with additional sensing modalities and verify their accuracy through correspondence algorithms~\cite{centerfusion, bevdepth}.

\begin{figure}[!t]
    \centering
    \vspace{0.15cm}
    \includegraphics[width=1.0\linewidth]{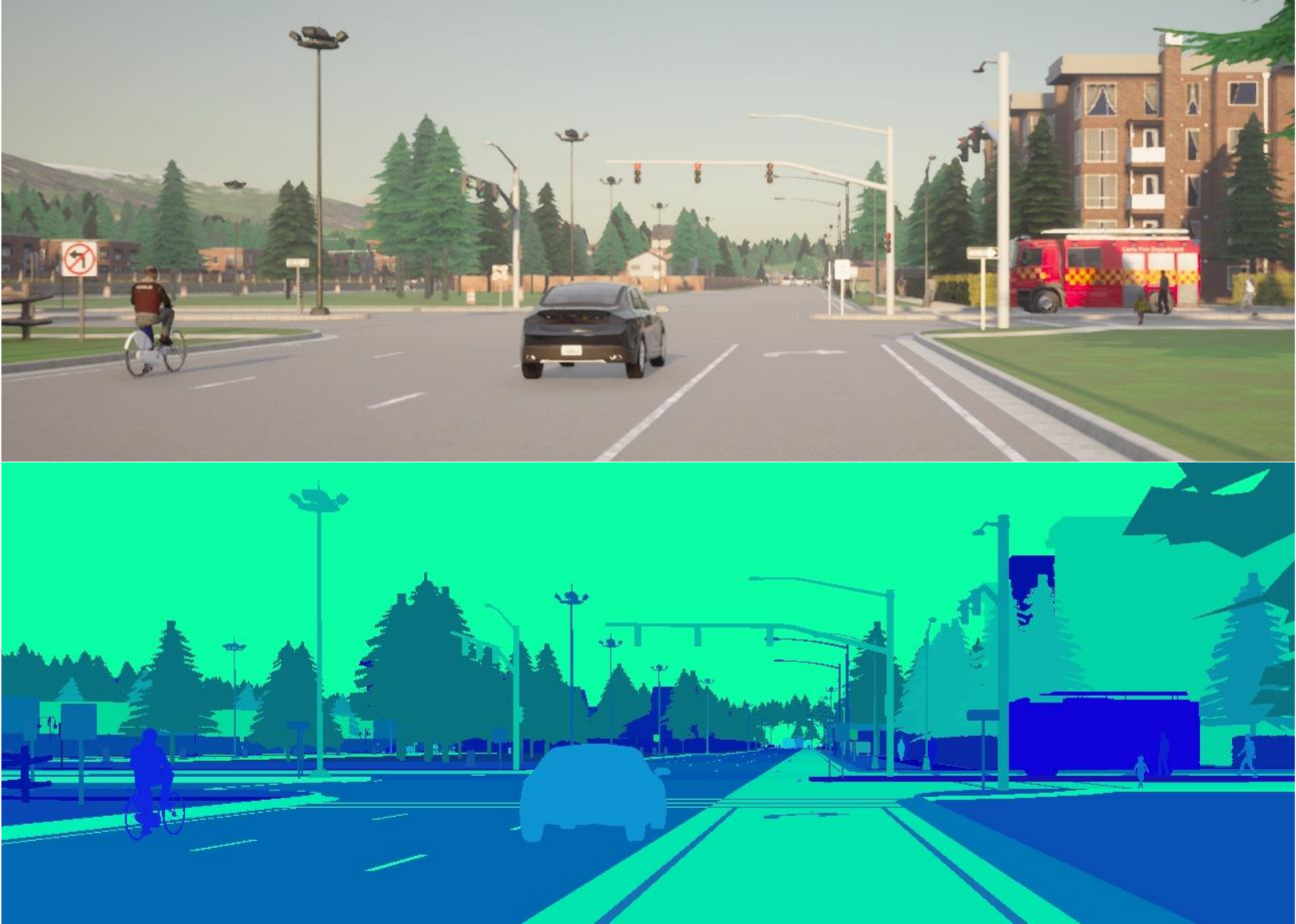}
    \caption{RGB Camera (top) and corresponding instance segmentation camera (bottom) views in CARLA Town06.}
    \label{fig:camera}
\end{figure}

\subsection{Ground Truth Annotation}

In nuScenes, annotations are created by human experts using specialized labeling tools, which entails significant cost and labor~\cite{nuScenesDevkitInstructions}. In CARLA, we can take advantage of privileged access to the ground truth information for all traffic participants, including their size, rotation, and translation. However, this also introduces a subtle but important complication: we cannot simply record the ground truth for every participant present in the simulation at a given time. Instead, it is necessary to precisely identify and record only those participants that are visible in at least one of the six camera views. Failing to filter for visibility would lead to significant false positive annotations, which could confuse the model and hinder the learning process.

Since the CARLA RGB camera does not provide information about which traffic participants are captured in its view, we employ an auxiliary instance segmentation camera that assigns a unique pixel value to every object in the scene. For each RGB camera, we mount a corresponding instance segmentation camera at the exact same position with identical calibration, ensuring perfectly aligned views (Fig.~\ref{fig:camera}). This allows us to accurately determine which traffic participants are visible in each camera view and record their corresponding information. Note that the process is used solely for generating ground truth annotations and is not utilized at any stage of model training or inference.

\begin{table*}[!t]
\renewcommand{\arraystretch}{1.2}
\caption{Summary of bird’s-eye-view (BEV) perception results for four models on the nuCarla \textit{validation} set, evaluated on seven training maps and six object classes using the official nuScenes detection metrics.}
\centering
\small
\begin{tabular}{p{7.0cm}|>{\centering\arraybackslash}p{1.0cm}>{\centering\arraybackslash}p{1.0cm}>{\centering\arraybackslash}p{1.0cm}>{\centering\arraybackslash}p{1.0cm}>{\centering\arraybackslash}p{1.0cm}>{\centering\arraybackslash}p{1.0cm}|>{\centering\arraybackslash}p{1.0cm}}

\hline
Methods & mAP\textuparrow & mATE\textdownarrow & mASE\textdownarrow & mAOE\textdownarrow & mAVE\textdownarrow & mAAE\textdownarrow & NDS\textuparrow\\
\hline
BEVFormer (Base)~\cite{bevformer, bevformerv2} & \textbf{0.813} & \textbf{0.266} & \textbf{0.063} & \textbf{0.053} & \textbf{0.728} & 0.177 & \textbf{0.778} \\
PETR (VovNet-GridMask-P4-1600x640)~\cite{petr, petrv2} & 0.745 & 0.438 & 0.088  & 0.061 & 0.906 & \textbf{0.137} & 0.710 \\
BEVDet (R50-4DLongTerm-Stereo-CBGS)~\cite{bevdet, bevdet4d} & 0.811 & 0.272 & 0.082 & 0.117 & 0.858 & 0.197 & 0.753 \\
FastBEV (R50-CBGS-4D)~\cite{fastbev} & 0.777 & 0.302 & 0.093 & 0.141 & 0.875 & 0.193 & 0.728 \\
\hline
\end{tabular}
\label{tab:summary_val}
\end{table*}

\begin{table*}[!t]
\renewcommand{\arraystretch}{1.2}
\caption{Summary of bird’s-eye-view (BEV) perception results for four models on the nuCarla \textit{test} set, evaluated on the two unseen test maps and six object classes using the official nuScenes detection metrics.}
\centering
\small
\begin{tabular}{p{7.0cm}|>{\centering\arraybackslash}p{1.0cm}>{\centering\arraybackslash}p{1.0cm}>{\centering\arraybackslash}p{1.0cm}>{\centering\arraybackslash}p{1.0cm}>{\centering\arraybackslash}p{1.0cm}>{\centering\arraybackslash}p{1.0cm}|>{\centering\arraybackslash}p{1.0cm}}

\hline
Methods & mAP\textuparrow & mATE\textdownarrow & mASE\textdownarrow & mAOE\textdownarrow & mAVE\textdownarrow & mAAE\textdownarrow & NDS\textuparrow\\
\hline
BEVFormer (Base)~\cite{bevformer, bevformerv2} & \textbf{0.579} & \textbf{0.451} & \textbf{0.068} & 0.173 & 1.006 & 0.208 & \textbf{0.599} \\
PETR (VovNet-GridMask-P4-1600x640)~\cite{petr, petrv2} & 0.514 & 0.596 & 0.098 & \textbf{0.144} & \textbf{0.852} & \textbf{0.186} & 0.569 \\
BEVDet (R50-4DLongTerm-Stereo-CBGS)~\cite{bevdet, bevdet4d} & 0.560 & 0.452 & 0.093 & 0.533 & 1.245 & 0.265 & 0.546 \\
FastBEV (R50-CBGS-4D)~\cite{fastbev} & 0.509 & 0.489 & 0.115 & 0.686 & 1.474 & 0.274 & 0.498 \\
\hline
\end{tabular}
\label{tab:summary_test}
\end{table*}

\subsection{Data Generation Limitations}

We discuss some limitations currently present in the data generation pipeline. In prebuilt CARLA maps, there exist parked vehicles and unattended bicycles or motorcycles that are embedded into the environment as static meshes rather than unique actors. Therefore, they do not appear as distinct instances in the segmentation cameras and lack accessible ground truth, resulting in missing annotations. To resolve this, we edit the CARLA source in Unreal Engine 4 to remove problematic static meshes, then recompile the CARLA package. This produces a custom build that differs from the official release, which may complicate custom data collection for other users.

\subsection{nuScenes Formatting}

We format our dataset in full alignment with nuScenes, following its naming conventions, annotation format, file hierarchy, and ensuring compatibility with the official Python API. Therefore, nuCarla offers a unique advantage over previous datasets~\cite{bench2drive, deepaccident, shift}: it provides a standardized closed-loop dataset to which any camera-based BEV perception model originally developed for nuScenes can be transferred without modification. We validate this interoperability by evaluating four state-of-the-art BEV perception models in the subsequent experiment section.
\section{Experiment}

\subsection{Model Configuration}
To thoroughly verify nuCarla, including the image quality of the RGB cameras, ground-truth annotation accuracy, and compatibility with nuScenes, we evaluated four state-of-the-art BEV perception models: BEVFormer~\cite{bevformer, bevformerv2}, PETR~\cite{petr, petrv2}, BEVDet~\cite{bevdet, bevdet4d}, and FastBEV~\cite{fastbev}. As our primary objective is to assess their interoperability with our dataset and their overall effectiveness in the closed-loop simulation environment, rather than systematically comparing performance or analyzing their architectural choices, we selected one variant from each model that offers a practical balance between efficiency and performance. Specifically, we used BEVFormer-Base, PETR-VovNet-GridMask-P4-1600x640, BEVDet-R50-4DLongTerm-Stereo-CBGS, and FastBEV-R50-CBGS-4D. For readers unfamiliar with these implementations, we refer to the original publications for further details on network implementation, training schedules, and resource utilization.

\subsection{MMDetection3D Upgrade}
We address practical compatibility concerns associated with existing E2E models~\cite{issue313, issue245, issue206}, which are built upon the MMDetection3D-1.0~\cite{mmdet3d} framework. This framework supports only earlier versions of PyTorch (\texttt{<}2.0) and CUDA (\texttt{<}12.0), limiting deployment on newer hardware such as NVIDIA H100 and GeForce RTX 50 series. While the upgraded MMDetection3D-2.0 framework has been released, migrating existing models would require extensive and potentially risky code refactoring.

To address this issue, we implemented a series of targeted patches to upgrade the MMDetection3D-1.0 framework for compatibility with the latest versions of PyTorch (2.7), CUDA (12.8), and modern GPUs. Our approach preserves the original model codebase, introducing only minimal changes necessary to resolve version conflicts. We validated the effectiveness of these modifications by successfully running the selected BEV models on both the official nuScenes dataset and our newly developed nuCarla dataset.

\subsection{Training and Evaluation Settings}

All models were trained from scratch on H100 GPUs for 24 epochs using the 700 training scenarios. Performance was evaluated on the 150 validation scenarios from the 7 trained maps, as well as the 150 test scenarios from the 2 unseen maps. The results were reported following the official nuScenes detection metrics over the six available classes, including average translation error (ATE), average scale error (ASE), average orientation error (AOE), average velocity error (AVE), and average attribute error (AAE). Mean Average Precision (mAP) and nuScenes Detection Score (NDS) were used to summarize overall performance.

\subsection{Validation Results}

Evaluation results of the four BEV models on the nuCarla validation set are shown in Table~\ref{tab:summary_val}. Overall, BEVFormer achieves the best results. Nevertheless, all models demonstrate robust performance, with mAP and NDS consistently exceeding 0.7. Note that since only a single variant from each model family was selected, these results are intended to demonstrate practical viability rather than provide a comprehensive benchmark.

Compared to their validation results on nuScenes, all models achieve substantially higher mAP and NDS on nuCarla, which we attribute to two primary factors. First, nuCarla excludes trailers and construction vehicles, which are infrequently annotated in nuScenes and exhibit relatively low performance (e.g. BEVFormer reports only 0.172 mAP for trailers and 0.129 mAP for construction vehicles). Their exclusion prevents these categories from adversely affecting the overall scores. Second, the reduced number of classes simplifies the problem, potentially lowering training complexity and facilitating rapid convergence.

\subsection{Test Results}

On the nuCarla test set (Table~\ref{tab:summary_test}), all models generalize reasonably well, despite performing worse than on the validation set. This is because the test set includes two previously unseen and practically more challenging maps. For example, Town10 is the flagship map of CARLA, featuring dense, vivid urban environments, whereas Mcity is a large open area with many traffic participants simultaneously in view. In contrast, on nuScenes, model performance on the validation and test sets does not differ significantly, suggesting that the distributions are similar. Overall, these results underscore the importance of evaluating model generalization on diverse and challenging scenarios.

\subsection{Visualization}

\begin{figure*}[!t]
    \centering
    \vspace{0.15cm}
    \includegraphics[width=1.0\linewidth]{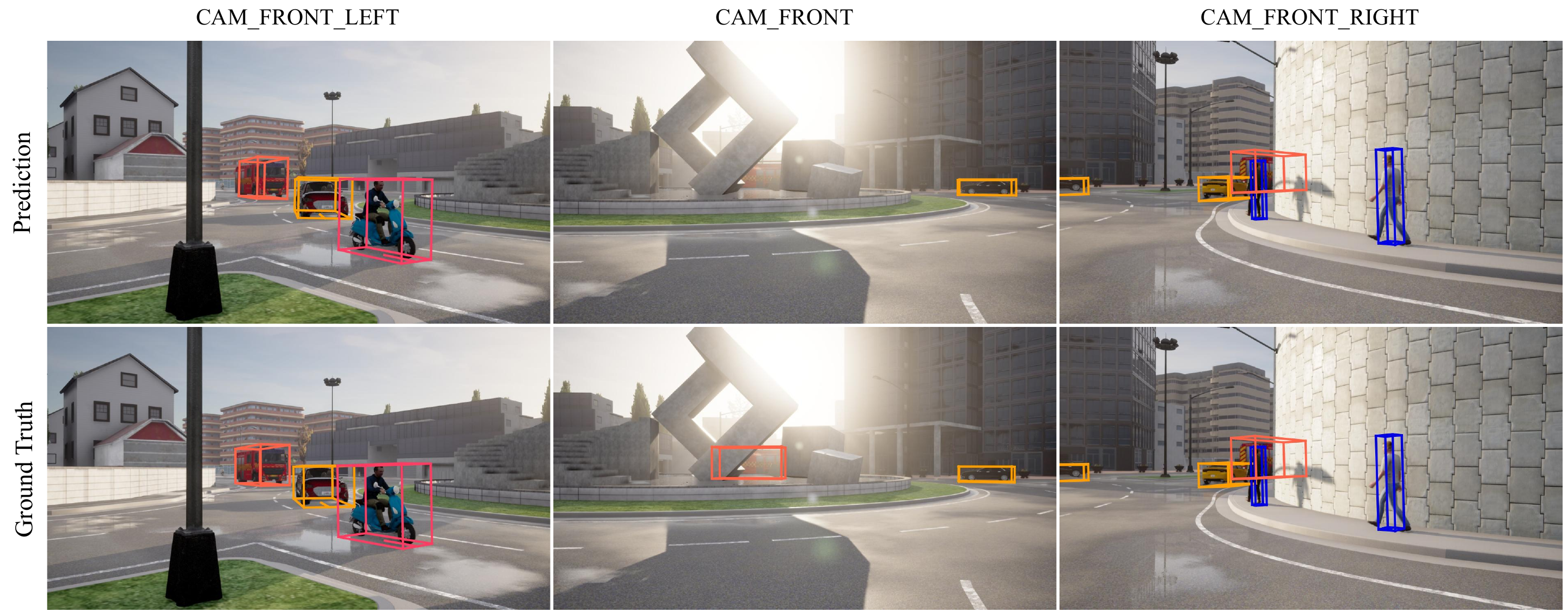}
    \caption{BEVFormer model predictions (top row) and ground truth annotations (bottom row) on a Town03 sample, featuring the front, front-left, and front-right camera views. 3D bounding boxes are colored according to the nuScenes colormap: cars (amber orange), trucks (coral red), pedestrians (blue), and cyclists (pink-red).}
    \label{fig:bev}
\end{figure*}

Fig.~\ref{fig:bev} shows a visual illustration of model predictions compared to ground truth. The predictions are generated by BEVFormer, evaluated on a sample from the Town03 map. To improve clarity, we display only the three front-facing cameras, rather than all six available views. As shown, the model predictions closely match the ground truth in translation, rotation, and size, accurately capturing all actors as indicated by the bounding boxes. The only missed prediction is a firetruck that is partially visible behind a statue on the far side of the roundabout. However, this object is also barely visible to humans due to backlighting, and its omission does not present any immediate safety concern.

\begin{table*}[!t]
\renewcommand{\arraystretch}{1.2}
\caption{Per-class detection metrics across six object classes, evaluated by BEVFormer, comparing nuCarla (on the left side of each cell) and nuScenes (on the right side) \textit{validation} set.}
\centering
\small
\begin{tabular}{p{1.4cm}|>{\centering\arraybackslash}p{2.2cm}>{\centering\arraybackslash}p{2.2cm}>{\centering\arraybackslash}p{2.2cm}>{\centering\arraybackslash}p{2.2cm}>{\centering\arraybackslash}p{2.2cm}>{\centering\arraybackslash}p{2.2cm}}
\hline
Class & AP\textuparrow & ATE\textdownarrow & ASE\textdownarrow & AOE\textdownarrow & AVE\textdownarrow & AAE\textdownarrow \\
\hline
Car & \textbf{0.792} \textbar\ 0.618 & \textbf{0.269} \textbar\ 0.462 & \textbf{0.098} \textbar\ 0.152 & \textbf{0.027} \textbar\ 0.067 & 0.896 \textbar\ \textbf{0.325} & 0.289 \textbar\ \textbf{0.196} \\
Truck & \textbf{0.828} \textbar\ 0.370  & \textbf{0.264} \textbar\ 0.726 & \textbf{0.042} \textbar\ 0.212 & \textbf{0.027} \textbar\ 0.093 & 0.691 \textbar\ \textbf{0.348} & 0.251 \textbar\ \textbf{0.193} \\
Bus & \textbf{0.826} \textbar\ 0.444 & \textbf{0.251} \textbar\ 0.753 & \textbf{0.035} \textbar\ 0.212 & \textbf{0.039} \textbar\ 0.099 & 0.918 \textbar\ \textbf{0.868} & \textbf{0.110} \textbar\ 0.270 \\
Pedestrian & \textbf{0.714} \textbar\ 0.494 & \textbf{0.359} \textbar\ 0.642 & \textbf{0.162} \textbar\ 0.295 & \textbf{0.157} \textbar\ 0.433 & 0.404 \textbar\ \textbf{0.360} & \textbf{0.001} \textbar\ 0.175 \\
Motorcycle & \textbf{0.849} \textbar\ 0.429 & \textbf{0.242} \textbar\ 0.639 & \textbf{0.028} \textbar\ 0.257 & \textbf{0.032} \textbar\ 0.438 & 0.914 \textbar\ \textbf{0.530} & \textbf{0.195} \textbar\ 0.265 \\
Bicycle & \textbf{0.867} \textbar\ 0.398 & \textbf{0.208} \textbar\ 0.554 & \textbf{0.011} \textbar\ 0.272 & \textbf{0.037} \textbar\ 0.484 & 0.547 \textbar\ \textbf{0.248} & 0.217 \textbar\ \textbf{0.024} \\
\hline
\end{tabular}
\label{tab:bevformer_nucarla}
\end{table*}

\subsection{Per-Class Metric Comparison}

To better assess the strengths and limitations of the dataset, Table~\ref{tab:bevformer_nucarla} provides a detailed per-class comparison between nuCarla (shown on the left side of each cell) and nuScenes (on the right side), using BEVFormer on the validation set. In nuScenes, the car category achieves notably higher Average Precision (AP) than the others, primarily due to the dominance of car annotations. In contrast, nuCarla exhibits more uniform scores, reflecting its balanced distribution.

For the Average Translation Error (ATE), Average Scale Error (ASE), and Average Orientation Error (AOE), varying degrees of improvement are also observed in nuCarla, particularly for underrepresented classes. While part of this improvement can be attributed to the more balanced class distribution, we believe another contributing factor is the inherently simpler and more structured nature of the CARLA environment. In this setting, traffic participants follow well-defined and consistent motion patterns, allowing models to effectively learn and generalize spatial relationships.

Conversely, the Average Velocity Error (AVE) is much higher in nuCarla. In nuScenes, the substantial proportion of stationary participants yields zero-velocity samples that are trivial to predict, thereby lowering the overall error. In contrast, all actors in nuCarla are actively traveling, which makes velocity estimation more challenging.

Finally, the Average Attribute Error (AAE) does not display a consistent pattern of discrepancies between the two datasets. This suggests that certain object classes within each dataset may be inherently easier to learn than others, such as pedestrians in nuCarla and bicycles in nuScenes. Additionally, some models demonstrate stronger attribute prediction capabilities; for example, PETR achieves better results in attribute estimation, even though its performance on other metrics is comparatively weaker.
\section{Conclusion}

In this work, we present nuCarla, a nuScenes-style bird’s-eye-view perception dataset designed for the CARLA simulator. To thoroughly validate nuCarla, we evaluate four state-of-the-art BEV perception models and provide pretrained weights. This facilitates the learning of meaningful intermediate representations and supports the advancement of end-to-end autonomous driving research through comprehensive closed-loop testing.

\vspace{0pt}
{
    \small
    \bibliographystyle{ieeenat_fullname}
    \bibliography{main}
}


\end{document}